\documentclass[letterpaper, 10 pt, conference]{ieeeconf}  

\IEEEoverridecommandlockouts                              

\usepackage{cite}
\usepackage{amsmath,amssymb,amsfonts}
\usepackage{algorithmic}
\usepackage{graphicx}
\usepackage{textcomp}
\usepackage{xcolor}
\def\BibTeX{{\rm B\kern-.05em{\sc i\kern-.025em b}\kern-.08em
T\kern-.1667em\lower.7ex\hbox{E}\kern-.125emX}}
\usepackage{graphics} 
\usepackage{epsfig} 
\usepackage{mathptmx} 
\usepackage{times} 
\usepackage{amsmath} 
\usepackage{inputenc}
\usepackage{siunitx}
\usepackage{nicefrac}
\usepackage{xcolor}
\usepackage{eqparbox}
\usepackage{tabularx}
\usepackage{booktabs}
\usepackage{multirow}
\usepackage{csquotes}
\usepackage{url}
\usepackage[caption=false,font=footnotesize]{subfig}
\usepackage{stmaryrd} 
\usepackage{makecell}
\usepackage[acronym, toc, shortcuts, nowarn]{glossaries}

\title{\LARGE \bf
A generic approach for reactive stateful mitigation of application failures in distributed robotics systems deployed with Kubernetes
}

\author{Florian Mirus$^{1, *}$, Frederik Pasch$^{1, *}$, Nikhil Singhal$^{1}$, and Kay-Ulrich Scholl $^{1}$
\thanks{$^{*}$ Equal contribution}
\thanks{$^{1}$Florian Mirus, Frederik Pasch, Nikhil Singhal and Kay-Ulrich Scholl are with Intel Labs, 
    Karlsruhe, Baden-W{\"{u}}rrtemberg, Germany
    {\tt\small \{florian.mirus, frederik.pasch, nikhil.singhal, kay-ulrich.scholl\}@intel.com}}
}

\makeglossaries
\newacronym{CV}{CV}{Computer Vision}
\newacronym{LIDAR}{LIDAR}{Light Detection and Ranging}
\newacronym{AMR}{AMR}{Autonomous Mobile Robot}
\newacronym{IoT}{IoT}{Internet of Things}
\newacronym{ROS2}{ROS2}{Robot Operating System}
\newacronym{ROS}{ROS}{Robot Operating System}
\newacronym{IMU}{IMU}{Inertial Measurement Unit}
\newacronym{AMCL}{AMCL}{Adaptive Monte Carlo Localization}
\newacronym{LSTM}{LSTM}{Long Short-Term Memory}
\newacronym{DDS}{DDS}{Data Distribution Service}
\newacronym{ASAM}{ASAM}{Association for Standardization of Automation and Measurement Systems}
\newacronym{KPI}{KPI}{Key Performance Indicator}
\newacronym{VM}{VM}{Virtual Machine}
\newacronym{CPS}{CPS}{Cyber-Physical System}
\newacronym{K8s}{K8s}{Kubernetes}
\newacronym{SERo}{SERo}{Scenario Execution for Robotics}

\begin{document}

\maketitle
\thispagestyle{empty}
\pagestyle{empty}

\begin{abstract}

Offloading computationally expensive algorithms to the edge or even cloud offers an attractive option to tackle limitations regarding on-board computational and energy resources of robotic systems.
In cloud-native applications deployed with the container management system \ac{K8s}, one key problem is ensuring resilience against various types of failures.
However, complex robotic systems interacting with the physical world pose a very specific set of challenges and requirements that are not yet covered by failure mitigation approaches from the cloud-native domain.
In this paper, we therefore propose a novel approach for robotic system monitoring and stateful, reactive failure mitigation for distributed robotic systems deployed using \acf{K8s} and the \ac{ROS2}. 
By employing the generic substrate of Behaviour Trees, our approach can be applied to any robotic workload and supports arbitrarily complex monitoring and failure mitigation strategies.
We demonstrate the effectiveness and application-agnosticism of our approach on two example applications, namely \ac{AMR} navigation and robotic manipulation in a simulated environment.

\end{abstract}

\section{INTRODUCTION}%
\label{sec:introduction}

Modern algorithms, particularly those employing cutting-edge AI, allow robots to reach a greater level of autonomy and fulfill more challenging tasks. 
However, on-board limitations regarding computational and energy resources are hindering factors regarding the deployment of such resource-hungry algorithms, particularly on mobile robots. 
On the other hand, the number and diversity of robots in a fleet for industrial automation will increase in the future. 
One attractive option to tackle both challenges is offloading most of the algorithmic workloads to the edge or even cloud to leverage massive computing power for robotics applications. 
However, given a large, heterogeneous fleet of robots with a diverse, challenging set of tasks and a complex stack of software deployed with a container management system such as \ac{K8s}, one key problem is how to mitigate failures and thereby minimize their impact on the robots’ and fleets’ task performance.
Potential failures in such a system range from failures of the compute nodes, over failures in the communication network to failures of the containerized applications themselves \cite{AbdollahiVayghan2019}.
In previous work, we already tackled temporary communication failures for a specific application use-case, namely \ac{AMR} navigation \cite{Mirus2024}.

In this paper, we focus on mitigation of application failures and investigate robotics-specific requirements, that are not yet covered by failure mitigation approaches from the cloud-native domain.
One core aspect of cloud-native applications is to ensure that failing micro-services are timely monitored and restarted such that, ideally, the end-user does not experience notable down-times.
However, in a cloud-native deployment, containers are mostly running isolated from the physical world whereas complex robotic systems interact with the physical world in real-time and stateful failure mitigation approaches from the cloud-native domain are not directly applicable.

\begin{figure}[t]
    \centering
    \includegraphics[width=1.\linewidth]{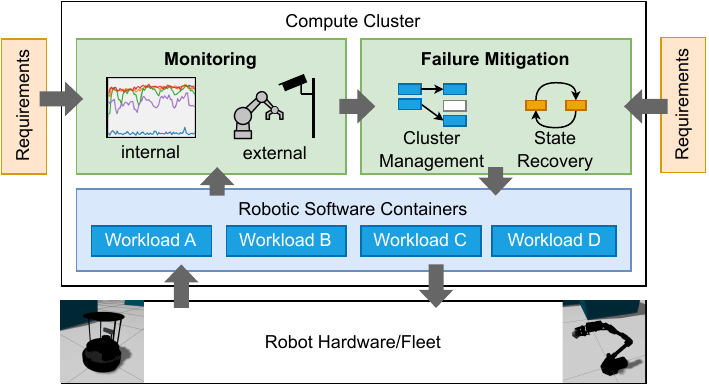}
    \caption{High-level overview of the monitoring and failure mitigation system.}
    \label{fig:overview}
    \vspace{-0.6cm}
\end{figure}

In this paper, we therefore propose a novel approach for robotic system monitoring and reactive failure mitigation for distributed robotic systems deployed using \acf{K8s} and \ac{ROS2}.
Our approach is agnostic to the application and considers the unique challenges of distributed robotics applications by introspectively monitoring robotics-specific metrics or, monitoring the overall system's behaviour through external sensors and applying robotic-specific failure mitigation.
One crucially important aspect is that our approach allows to preserve the last healthy state of the robotic system before the failure occurred and transfer it once the failure is resolved.
This enables the robotic application to continue its task at the point where the failure of the workload occurred.
Furthermore, our approach offers different methods for handling application failures varying in the time necessary for bringing the workload back to a functional state as well as the computational resources necessary.
Finally, our approach employs Behaviour Trees \cite{Colledanchise2018} to encapsulate the control of the failure mitigation and thus allows arbitrarily complex failure mitigation strategies.
We demonstrate the effectiveness and application-agnosticism of our approach on two example applications, namely \ac{AMR} navigation and robotic manipulation, in a simulated environment.

In summary, our three main contributions are: 1) an application-agnostic, reactive failure mitigation system based on the generic substrate of Behaviour Trees for distributed robotic systems deployed using \acl{K8s} and \ac{ROS2} allowing to preserve the last healthy state prior to the failure. 
2) a robotics-specific workload monitoring system to detect failures either trough introspection, i.e., monitoring system diagnostics and \acp{KPI}, and/or through external supervision, i.e., observing the overall system's behaviour with external sensors and comparing the expected with actually observed behaviour.
3) several recovery strategies resulting in a trade-off between system-downtime and demand for computational resources.

\section{RELATED WORK}%
\label{sec:related_work}

Offloading robotic software to the edge/cloud offers a powerful enhancement to various robotic tasks \cite{Kehoe2015} by leveraging high computing power and large storage spaces in the robotics domain.
In recent years, researchers proposed a plethora of innovative approaches and concepts for employing edge-computing in the robotics domain ranging from applications such as perception \cite{Beksi2015, Rashid2024}, grasping \cite{Kehoe2014, Zahid2024}, motion planning \cite{Ichnowski2020} and mobile navigation \cite{Groshev2022, Balogh2021}.
Offloading workloads to the edge or even cloud is particularly useful for resource-hungry algorithms \cite{Chinchali2021}, for instance, based on modern machine learning approaches \cite{Rashid2024}.
Chen et.~al. \cite{Chen2024} proposed a strategy for optimal service deployment in the cloud to ensure that the strict requirements of latency-critical robotics applications  are met.

These scientific advancements are complemented from a system perspective by several architectural solutions, which aim to simplify the access to cloud computing for roboticists.
For instance, FogROS2 \cite{Ichnowski2023} automatically provisions a cloud computer to deploy and launch \ac{ROS2} nodes while KubeROS \cite{Zhang2023} uses \ac{K8s} to facilitate the deployment of \ac{ROS2}-based applications across robot, edge, and cloud. 

One key problem when deploying containerized applications using a container management system such as \ac{K8s} is ensuring resilience against various types of failures \cite{AbdollahiVayghan2019}, which can range from communication network failures, over compute node failures to failure of the applications or pod processes themselves. 
In the context of \ac{AMR} navigation, a strategy to tackle temporary communication network failures by leaving minimal fallback workloads on the robot's on-board compute has been proposed in \cite{Mirus2024}.
In cloud-native applications using \ac{K8s} to deploy containers, failure mitigation approaches are typically separated in two categories, namely reactive and proactive \cite{Huang2020}.
Reactive approaches re-launch a failed service after a fault has occurred and was detected by the system.
Proactive approaches on the other hand aim to predict potential failures before they happen employing, e.g., learning approaches ranging from forecasting neural networks such as \acp{LSTM}, k-means, or reinforcement learning and take countermeasures accordingly.
Another important aspect is preserving the state of the failed application.
In cloud-native applications, there is a large variety of approaches for stateful container migration \cite{Souza2020} and stateful failure mitigation \cite{Tran2022}. 
Other approaches focus on mitigating failures occurring in the container management system \ac{K8s} itself \cite{Barletta2024}.

In this paper, we aim to close the gap for a robotics-specific stateful failure mitigation for distributed robotics applications deployed with \ac{K8s}.
Our focus lies on a reactive approach detecting application failures either through introspection of relevant \acp{KPI} or external supervision, and a generic failure mitigation method based on Behaviour Trees.
The introspection monitoring draws inspiration from monitor templates for robotic systems proposed in \cite{Jiang2017} but, for simplicity, only uses user-defined monitors instead of automatically generated ones as proposed in \cite{Jiang2017}.
To the best of our knowledge, our work is the first to consider robotics-specific requirements for stateful mitigation of application failures within robotic systems deployed with the container management system \acl{K8s}.

\section{STATEFUL FAILURE MITIGATION}%
\label{sec:stateful_failure_mitigation}

\begin{figure}
    \centering
    \includegraphics[width=0.8\linewidth]{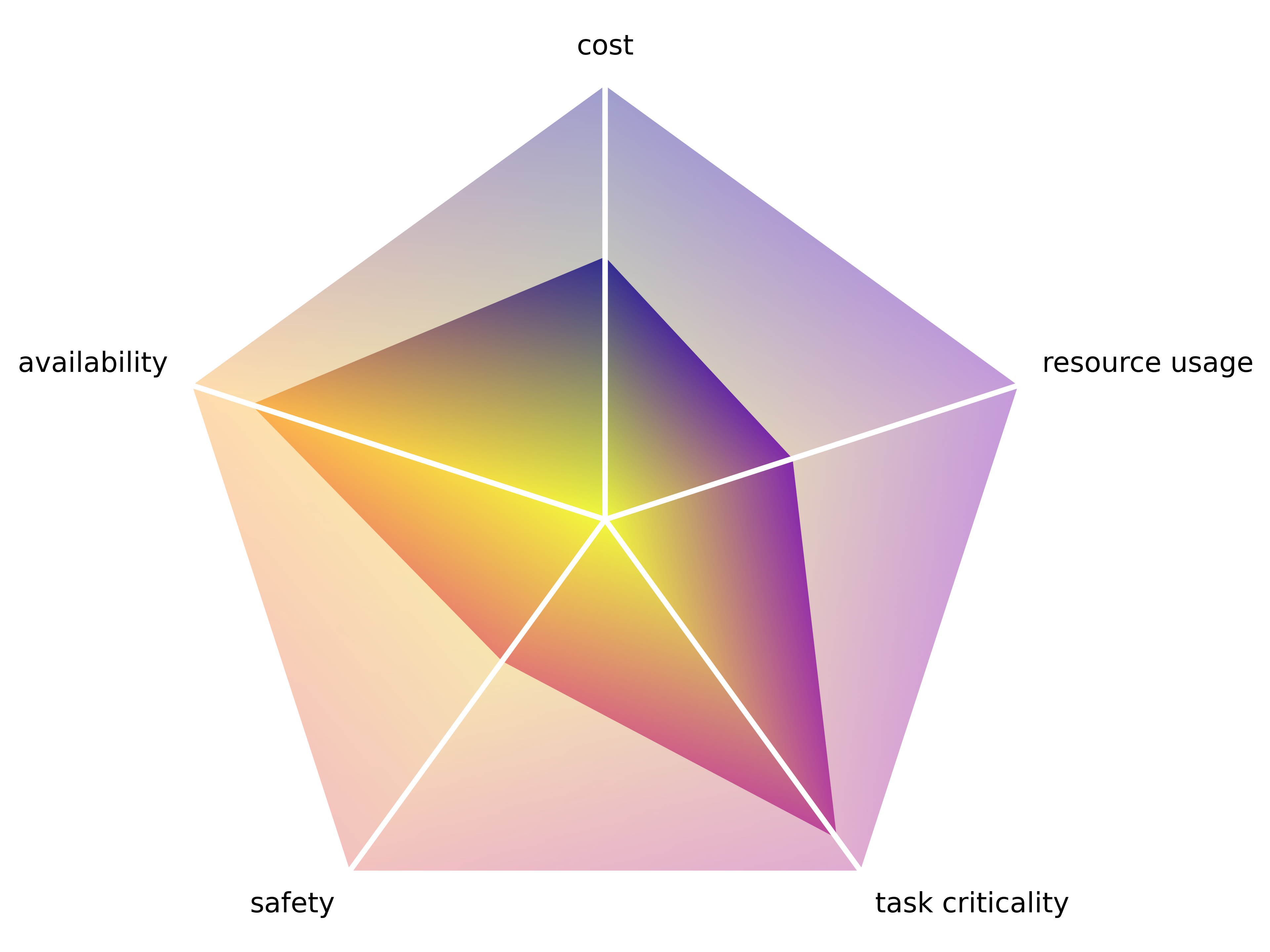}
	\caption{Factors and possible weights for selection of Failure Monitoring and Mitigation Strategy.}
    \label{figtaxonomy}
    \vspace{-0.5cm}
\end{figure}

Our reactive, stateful failure mitigation system consists of two main components: the monitoring system for failure detection as well as the actual failure mitigation for bringing the workload back to a functional state while preserving the last healthy state prior to the failure (see Fig~\ref{fig:overview}).
Both components rely on Behaviour Trees \cite{Colledanchise2018} to describe and apply the monitoring and failure mitigation procedures during execution time. 
Behavior trees are a powerful and generic substrate, that provides a formal and extensible structure for the monitoring and failure mitigation logic, which can be used to create arbitrarily complex monitoring systems and failure mitigation procedures while being human-readable and verifiable. 
Importantly, our failure mitigation approach is executed next to the main application without requiring to change the application's source code.

\subsubsection{Taxonomy}
\label{ssubsec:taxonomy}

\begin{figure}[t]
    \centering
    \includegraphics[width=1.\linewidth]{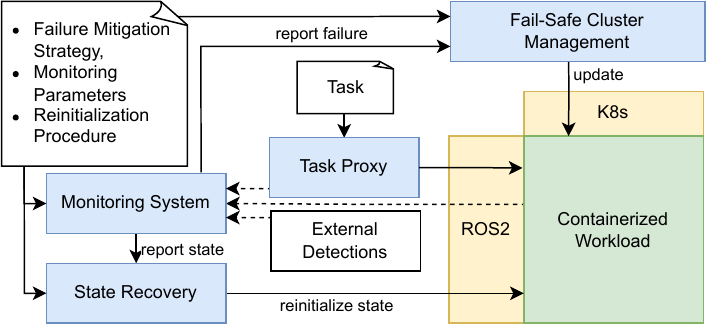}
    \caption{System Architecture}
    \label{fig:architecture}
    \vspace{-0.6cm}
\end{figure}

Monitoring advanced robotic systems can be very difficult due to the systems' inherent complexity \cite{Jiang2017}.
In general, there is plethora of levels at which robotics systems can and should be monitored to detect failures in the hardware, the software, at application level, at behaviour level or from a safety perspective. 
The choice of both, the monitoring requirements and the actual failure mitigation strategy could be tailored depending on the requirements of the application or task at hand.
For instance, if the robot is performing a difficult task handling expensive materials that could be damaged while a failure occurs, it is more important to keep the system's down-time to a minimum at the cost of additional computational resources.
For other cases, some level of system-downtime could be acceptable while stricter requirements regarding computational resources apply.
Therefore, we propose to select the focus of the monitoring and failure mitigation system according to a taxonomy as depicted in Fig.~\ref{figtaxonomy} considering the requirements (such as safety, availability, resource usage, safety, task criticality just to name a few options) of tasks the robotic system is performing.

\subsection{Monitoring System}
\label{subsec:monitoring}

Application- and workload-monitoring is one key ingredient to ensure meaningful and functional operation of edge- and/or cloud-systems. 
However, monitoring a complex robotic system interacting with the physical world poses unique challenges compared to cloud-native applications.
In this paper, we focus concretely on two main monitoring approaches, namely introspective monitoring of software workloads and external supervision of the robot's behaviour and task performance.

\subsubsection{Introspective monitoring}
\label{ssubsec:introspective_monitoring}

\begin{figure}
    \centering
    \includegraphics[width=\linewidth]{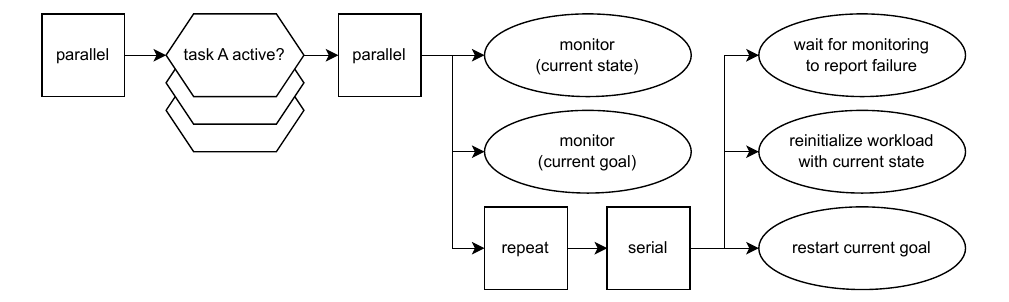}
	\caption{Example Behaviour Tree to apply basic failure mitigation for a task and an automatically restarting workload (e.g., using Kubernetes deployment)}
    \label{fig:example_bt}
   \vspace{-0.4cm}
\end{figure}

Here, we mainly apply \textit{time-related monitors} according to the templates proposed in \cite{Jiang2017}, which are characterized by a \acs{ROS2} topic's frequency.
However, availability and frequency of a certain topic depend on the robot's current task and therefore a more generic monitoring solution on top of the topic monitors is necessary.
For instance, consider an \ac{AMR} equipped with a robotic arm, i.e., a mobile manipulator: if the frequency of the velocity command sent to the actuators of the mobile base is not consistently meeting a desired level while the robot is supposed to be navigating towards a desired goal position, the system is most likely in a failure state.
In contrast, if there are no velocity commands sent to the mobile base, while the robot is standing to manipulate objects with its arm and/or end-effector, the system is considered healthy if the joint-states of the arm are delivering their data at the expected frequency. 
At the same time, the availability of meaningful data from various sensor sources such as \ac{LIDAR}, \ac{IMU}, Cameras or torque sensors, needs to be monitored permanently in parallel to the task-dependent monitoring objectives, which are only active if certain conditions are fulfilled.
For being able to model such hierarchical and situational monitoring requirements, we employ the generic substrate of Behaviour Trees \cite{Colledanchise2018}.
Fig.~\ref{fig:example_bt} visualizes one possible Behaviour Tree for the aforementioned mobile manipulator example.

\subsubsection{External supervision}
\label{ssubsec:external_monitoring}

Our second monitoring approach is unique for \acp{CPS}.
As robotic systems interact with the physical (or simulated) world, the resulting behaviours (e.g., movement of the manipulator or the mobile base driving along path) can be monitored through external sensors such as cameras or \acp{LIDAR} and state-of-the-art object detection and tracking approaches.
The essential part of this monitoring approach is to compare the behaviour observed with the external sensor(s) with the expected behaviour, which is known to the system through introspection.
If this comparison yields a significant discrepancy between observed and expected behaviour, the failure mitigation component of our system needs to apply counter-measures to bring the system back to a functional state.

\subsection{Failure Mitigation}
\label{subsec:failure_mit}

In case the monitoring system as described in Sec.~\ref{subsec:monitoring} reports a critical failure of a workload, the failure mitigation is responsible for bringing the workload back to a functional, healthy state.
Our system architecture, the interplay between the two main components and how they are connected to the monitored workload is visualized in Fig.~\ref{fig:architecture}.
Similar to our monitoring approach, the failure mitigation procedure is encapsulated in the generic substrate of Behaviour Trees to, in general, allow the system to perform any failure mitigation procedure.
Concretely, we focus on four classes of failure mitigation strategies, which result in a trade-off between down-time of the overall system after the failure occurred and the additional computational resources required for the mitigation as depicted in Fig.~\ref{fig:fail_mit_tradeoff}.

\subsubsection{Failure mitigation strategies}
\label{ssubsec:fail_mitigation_strategies}

\begin{figure}
    \centering
    \includegraphics[width=\linewidth]{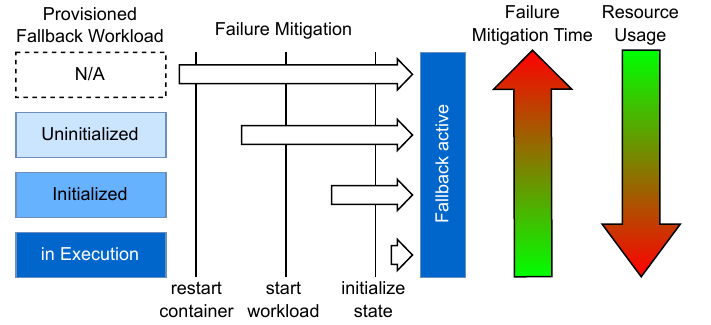}
	\caption{Required steps for the different failure mitigation strategies and their trade-off in terms of failure mitigation time and resource usage. Depending on the provisioned fallback workload, the steps of restarting the container, workload startup and state initialization need to be executed.}
    \label{fig:fail_mit_tradeoff}
    \vspace{-0.5cm}
\end{figure}

The simplest way is to restart the failing workload from scratch. 
However, this is the slowest recovery strategy as it requires both failure mitigation at the level of the \acl{K8s} cluster as well as re-initialization of the failed application and thus results in some down-time of the workload.
On the other hand, this recovery strategy does not require additional computational resources except for the monitoring during operation. 
Another option is to run a fallback instance in parallel to the main workload to speed up the recovery time.
The total time necessary for recovery depends on the state the fallback workload is in, which could be either of 1) uninitialized 2) initialized and 3) in execution while receiving external data but not sending data to the remainder of the system (see Fig.~\ref{fig:fail_mit_tradeoff}). 

All four strategies take some time at \ac{K8s} cluster level either for restarting the failing workload or connecting the rest of the system to the fallback workload.
Furthermore, the time for recovery at application level varies depending on the state of the fallback instance: if the fallback workload is only started, it needs to be initialized, possibly handed over the last healthy state of the failed main workload and pick up the task where the main workload failed.
If the fallback workload is already initialized or even in execution mode, the necessary time for recovery reduces accordingly (see also Fig.~\ref{fig:fail_mit_tradeoff} and~\ref{fig:failure_mitigation_timing_nav2} and Eq.~\ref{eq:recovery_time}).

\subsection{State Recovery}

In addition to re-launching the failing workload or connecting the running system to the fallback workload, it is crucially important that the new instance (either fallback or restarted) properly picks up where the main workload failed, i.e., the last healthy state needs to be stored during operation and handed over from the failing workload. 
Depending on the recovery strategy employed as described in Sec.~\ref{ssubsec:fail_mitigation_strategies}, the steps performed during the recovery procedure potentially differ. 
Therefore, the proposed failure mitigation is able to run arbitrarily complex recovery procedures encoded as Behaviour Tree, which it receives as a user-defined description (marked as \textit{requirements} in Fig.~\ref{fig:overview}). 
Additionally, it is possible that the failing workload critically depends on another (healthy) workload.
In such a case, it could even be necessary to restart/recover the healthy workload as well. 
The proposed recovery approach based on Behaviour Trees also supports such complex recovery scenarios by encapsulating dependencies within the Behaviour Tree description.

\subsection{Implementation Details}
\label{subsec:implementation_details}

To realize the failure mitigation strategies described in Sec.~\ref{ssubsec:fail_mitigation_strategies}, we rely on off-the-shelf \acf{K8s} components: the restart from scratch is realized through a \ac{K8s} deployment \cite{K8sDeployment} while rewiring the system to connect to the fallback workloads instead of the failed workloads is based on dynamically changing \ac{K8s} network policies \cite{K8sNetworkPolicies}.
For the realizing the Behaviour Tree representation and execution of the workload monitoring and failure mitigation, we employ \acl{SERo} \cite{Pasch2024, SERoCode}, a software library that translates scenarios or, in our case and more generally, Behaviour Tree description files written in the OpenSCENARIO~2 \cite{OpenScenario} language to a Python Behaviour Tree \cite{Pytrees} and executes it. To keep the high-level task request active, we establish a task proxy, that is capable of handling a possible workload re-initialization.

\section{EXPERIMENTS}%
\label{sec:experiments}


\subsection{Experimental Setup}%
\label{subsec:experimental_setup}

We demonstrate the effectiveness of our reactive failure mitigation system for robotics applications on two example applications, namely \ac{AMR} navigation and robotic manipulation in the Gazebo simulator \cite{Koenig2004, Gazebo}.
We use two different domains to demonstrate that our approach is agnostic of the application or workload it is monitoring and protecting.
For \ac{AMR} navigation, we use the Turlebot 4 \cite{Turtlebot4} as robot platform and the WidowX-200\cite{WX-200} robotic arm for manipulation.
We containerize the \ac{ROS2} frameworks Nav2 \cite{Macenski2023, Nav2} and MoveIt2 \cite{Coleman2014, MoveIt2} for navigation and robotic manipulation respectively and execute one simple task per example application.
For both applications, we execute a basic task, i.e., move the robot platform or arm to a series of user-defined goal positions, and after a user-defined time time $t_{failure}$, we delete the \ac{K8s} pod containing the main workload (either Nav2 or MoveIt2) to artificially inject a failure.
Again, we use \acl{SERo} \cite{Pasch2024, SERoCode} to define and control the high-level scenarios.
In both cases, our monitor checks the frequency of specific topics, i.e., the velocity commands sent to mobile base for Nav2 or the joint states for the robotic arm, and reports a failure once the stream of messages gets interrupted due to the pod containing the main workload being deleted.
To demonstrate the external behaviour monitoring, we only consider the \ac{AMR} navigation use-case: here, our monitor compares the expected velocity command received from introspection with the velocity command calculated from external observations. 
These observations are obtained by placing a camera nearby and detecting the poses of ArUco markers attached to the robot over time. 
To inject a failure, we remap the velocity commands such that the system assumes the mobile base receives meaningful velocity commands, whereas it actually stops moving.

\subsubsection{Metrics}
\label{ssubsec:metrics}

The first metric to consider is the time necessary to recover from an application failure and to bring the overall robotic system back to a functional state. 
This recovery time $t_{recovery}$ is the sum of 1) the time $t_{detection}$ necessary to detect the failure 2) failure mitigation time at cluster level (microservice/pod restart and/or time for adjusting network connections) and 3) restart and 4) re-initialization time at application level, i.e., 

\begin{equation}
    \label{eq:recovery_time}
    t_{recovery} = t_{detection} + t_{cluster} + t_{startup} + t_{re-initialization}.
\end{equation}

\begin{figure}[t]
	\vspace{1.5mm}
	\centering
	\includegraphics[width=\linewidth]{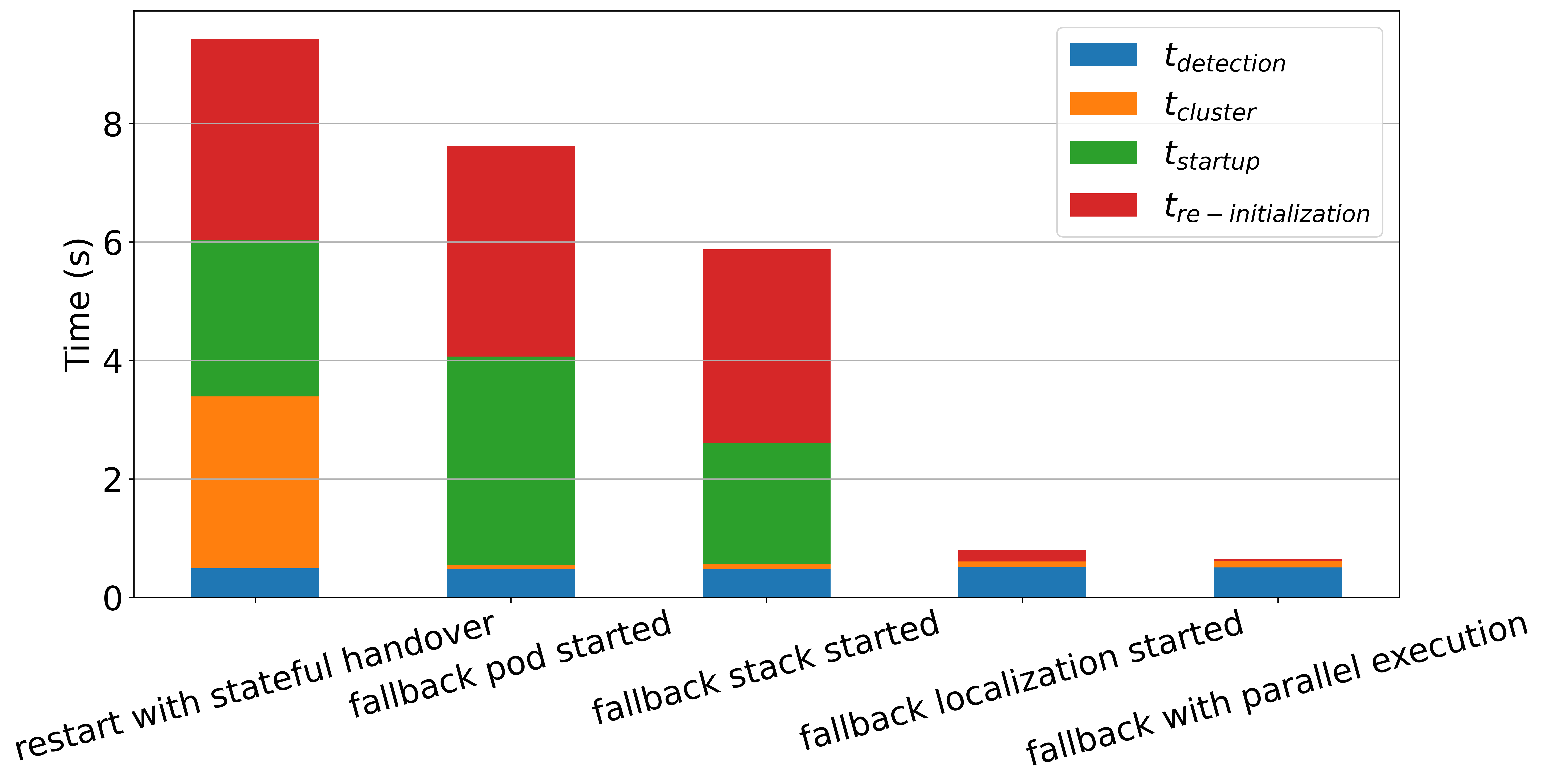}
	\caption{Evaluation of failure recovery approaches regarding necessary time for failure mitigation.}
	\label{fig:failure_mitigation_timing_nav2}
 \vspace{-0.5cm}
\end{figure}

The second metric is the usage of computational resources.
We measure the CPU usage $\gamma_{c,t_{i}}$ at timestamp $t_{i}$ for $i \in \{1, \ldots, n\}$ for each container $c$ separately over the duration of an entire experimental run.
Therefore, we use CAdvisor \cite{CAdvisor}, a tool for analyzing and exposing resource usage and performance data from running containers.
The unit for measuring CPU usage is the \ac{K8s} standard milliCPU \cite{K8sCPU} per second.
The final metric is the total average CPU usage of the cluster $\sigma(CPU)$, i.e., the sum of the mean CPU usage $\sigma(c)$ per container $c$  for all containers in the cluster $\mathfrak{C}$:

\begin{equation}
    \label{eg:mean_cpu_usage}
    \sigma(CPU) = \sum_{c \in \mathfrak{C}} \underbrace{\frac{1}{n} \sum_{i=1}^{n} \gamma_{c,t_{i}}}_{=\sigma(c)}.
\end{equation}

\subsection{Evaluation of recovery strategies}
\label{subsec:recovery_eval}

\subsubsection{\ac{AMR} navigation use-case}

Fig.~\ref{fig:failure_mitigation_timing_nav2} visualizes the individual parts of the recovery time $t_{recovery}$ according to Eq.~\ref{eq:recovery_time} for each recovery strategy.
Note that we only compare our recovery strategies against each other without a baseline, as there is - to the best of our knowledge - no other framework (e.g., FogROS2 \cite{Ichnowski2023} and KubeROS \cite{Zhang2023}), that offers the functionality we are proposing in this paper.
The closest off-the-shelf \acl{K8s} counterpart is Stateful Sets \cite{K8sStatefulSets}, which however would require adaptations to the application itself, whereas our solution just uses off-the-shelf \ac{ROS2} and \ac{K8s} tools without requiring application changes.

As Nav2 consists of several components, we investigate different levels of initialization for the recovery strategies involving a fallback workload ranging from just the \ac{K8s} pod being started, over the Nav2 stack being started, to the fallback container running the localization workload and even the full Navigation stack in parallel.
The necessary time $t_{detection}$ to detect the failure depends on the requirements specified by the user and is set to \SI{500}{\milli\second} in our experiments for all recovery strategies.
The failure mitigation time at cluster level $t_{cluster}$ shows significant differences among recovery strategies: restarting a container from scratch takes \ac{K8s} significantly longer (on average \num{32} times longer) than patching network policies to connect the system to the fallback variant of the failing workload (\SI{2.9}{\second} vs. \SI{0.1}{\second} on average).
Furthermore, with increasing initialization of the fallback workload, the startup time $t_{startup}$ and re-initialization time $t_{reinitialization}$ (green and red bars in Fig.~\ref{fig:failure_mitigation_timing_nav2} respectively) decrease proportionally. 
For the fallback workload in full execution mode, application startup is not needed and hence, the startup time collapses to $t_{startup}=0$.

\begin{figure}[t]
	\vspace{1.5mm}
	\centering
    \includegraphics[width=\linewidth]{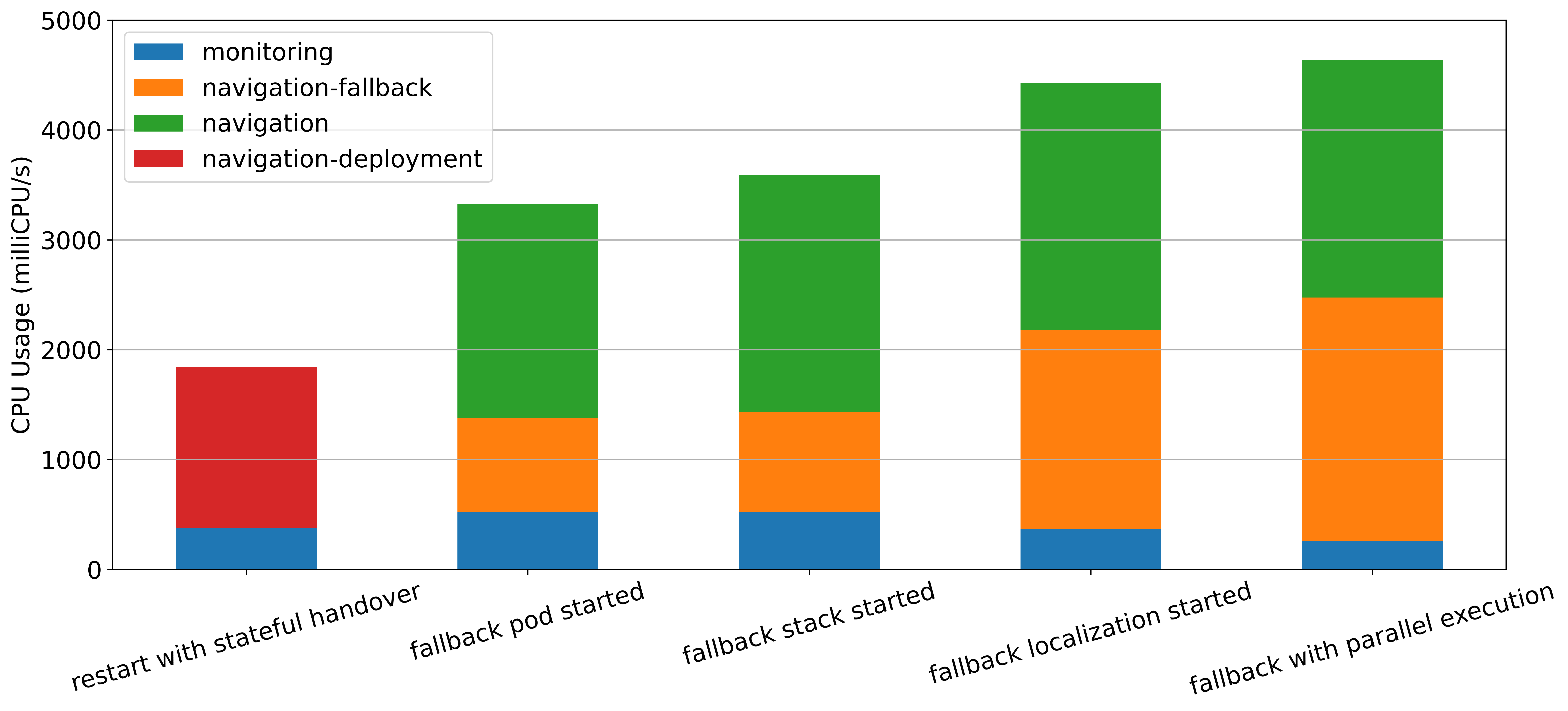}
	\caption{Experimental evaluation of failure recovery approaches regarding CPU load.}
	\label{fig:failure_mitigation_cpu_nav2}
 \vspace{-0.5cm}
\end{figure}

Fig.~\ref{fig:failure_mitigation_cpu_nav2} visualizes the total average CPU usage $\sigma(CPU)$ for each recovery strategy with the stacked colored bars representing the mean CPU usage of individual \ac{K8s} pods containing either the workload instances, i.e.,  Nav2, or our monitoring and failure mitigation system.
As expected, the failure mitigation strategy starting the failing workload from scratch (leftmost bar in Fig.~\ref{fig:failure_mitigation_cpu_nav2}) consumes the least resources on average, as there is no fallback container running in parallel to the main workload.
The recovery strategies that hold an uninitialized fallback variant of the navigation stack in parallel to the main system consume at least the same resources as the restart from scratch in addition to the resources needed for running the uninitialized fallback workload in parallel.
The uninitialized container consumes around \SI{60}{\percent} of the resources necessary for the main navigation container.
Similarly, the fully initialized fallback navigation stack running in execution mode in parallel consumes the same resources as the main workload. 
Hence, these recovery strategies are the most-resource hungry among all recovery strategies and additionally require one fallback instance for each monitored workload whereas an uninitialized workload can serve as a fallback for several main workload instances (e.g., in a robot fleet).
The additional resources necessary for the monitoring component of our system is (almost) identical for all recovery strategies.

\subsubsection{Manipulation use-case}
 
\begin{figure}[t]
	\vspace{1.5mm}
	\centering
	\includegraphics[width=\linewidth]{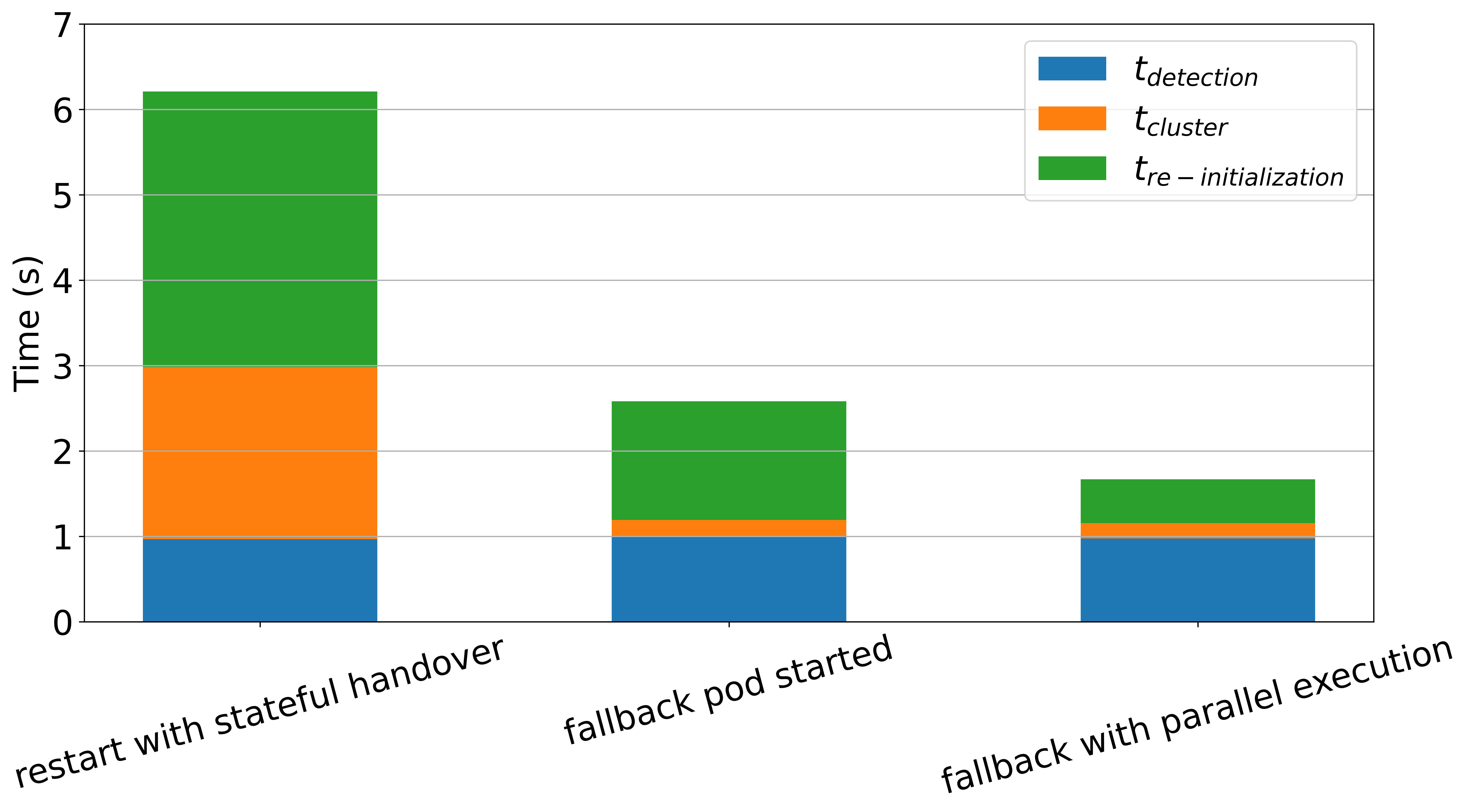}
	\caption{Navigation: Experimental evaluation of failure recovery approaches regarding necessary time for failure mitigation.}
	\label{fig:failure_mitigation_timing_moveit2}
 \vspace{-0.5cm}
\end{figure}

Fig.~\ref{fig:failure_mitigation_timing_moveit2} visualizes the recovery time for each failure mitigation strategy.
Due to application differences, only a subset of the mitigation strategies from the navigation use case is applicable for manipulation, namely restart from scratch, a fallback workload with just the \ac{K8s} being started and a fallback workload in full execution mode.
As expected, we observe tendencies similar to the \ac{AMR} navigation use-case: starting the failing workload from scratch is the slowest whereas parallel execution offers the fastest recovery.
The resource usage of the recovery strategies for manipulation is similar to the navigation example, hence we omitted a dedicated figure as it offers little additional information over Fig.~\ref{fig:failure_mitigation_cpu_nav2}.

\subsection{Scaling considerations}
\label{subsec:scaling}
In Sec.~\ref{subsec:recovery_eval}, we have seen that there is indeed a trade-off between recovery time and necessary computational resources regarding the choice of the failure mitigation strategy.
Depending on the task and application of the robotic system, the choice for a suitable recovery strategy might vary. 
For instance, time-critical tasks involving expensive goods that call for as little down-time as possible and hence are more relaxed regarding additional resource usage will employ a recovery strategy with a fallback workload running in execution mode in parallel to the main workload.
On the other extreme, workloads running in a resource-constrained cluster with less strict requirements regarding system-downtime will employ the recovery strategy starting the failing workload from scratch.

Consider a fleet of $N$ robots and, for simplicity, only one workload per robot to monitor (e.g., navigation) with a failure of this workload causing a system downtime of $t_{S}$ depending on the recovery strategy $S$.
Additionally, let $I = \{t_{1}, \ldots, t_{N}\}$ be a discrete time interval of interest and $X$ the number of expected failures of the entire fleet within an interval of length $N$.
To ensure that the expected down-time of the system does not exceed the down-time $t_{S}$ of the selected recovery strategy $S$, we need to run a sufficient number of fallback workloads.
Sufficient means that the probability that more failures than available fallback workloads occur within a time-window $D_{k} \subset I$ of length $t_{S}$ for $k = 1, \ldots, N-t_{S}+1$  is sufficiently low.
Assuming that the probability of a failure occurring at time-step $t_{i}$ within the interval $I$ is equal for all $i=1, \ldots, N$, the number of possible failure occurrences $f_{1}, \ldots f_{X}$ during the time interval $I$ is lower than
\begin{equation}
    \label{eq:binom}
    \binom{N+X-1}{N-1},
\end{equation}
because we only count instances with $f_{j} \in D_{k}$ for all $j=1, \ldots, X$ for one time-window $D_{k}$.
For example, let's assume a (fairly high) failure rate of \num{1} failure per hour per robot for a fleet of $N=1000$ robots, i.e., a total of \num{1000} failures per hour.
Hence, in a time interval $I$ of \SI{30}{\second}, $\frac{1000}{120}\approx8.3$ failures occur.
By employing \num{4} fallback instances, the probability of the $8.3 - 4 \approx 5$ remaining failures to occur within a time-window $D_{k}$ of length $t_{S} = \SI{6}{\second}$ (which is roughly the down-time we measured for navigation for the recovery strategy holding an uninitialized fallback workload), i.e., leading to a longer system-downtime is at \SI{1.2}{\percent}.
To reduce the down-time by factor \num{10} compared to the restart from scratch by having one fallback workload in parallel execution mode for all \num{1000} robots, the CPU usage in turn would be double whereas \num{4} additional fallback workloads would reduce the down-time by a factor of \num{2} and only add \SI{0.4}{\percent} resource usage compared to the restart from scratch.
Hence, a recovery strategy holding an uninitialized fallback workload in parallel offers a good balance between system down-time and additional resource usage.
Particularly, for larger robot fleets, the additional compute resources necessary in such a setting is neglectable.

\section{DISCUSSION}%
\label{sec:discussion}


\subsection{Conclusion}%
\label{subsec:conclusion}
In this paper, we presented an application-agnostic, reactive failure mitigation system based on the generic substrate of Behaviour Trees for distributed robotic systems deployed using \acl{K8s} and \ac{ROS2} allowing to preserve the last healthy state prior to the failure.
Our approach consists of a robotics-specific workload monitoring system to detect failures either trough introspection, i.e., monitoring system diagnostics and \acp{KPI}, and/or through external supervision, i.e., observing the overall system's behaviour with external sensors.
Finally, we presented different recovery strategies resulting in a trade-off between system-downtime and demand for computational resources and demonstrated the effectiveness of our approach at two example applications, namely \ac{AMR} navigation and robot manipulation.

\subsection{Future Work}%
\label{subsec:future_work}

Although we believe that our approach is an important step towards making distributed robotic systems deployed with a container management system such as \ac{K8s} resilient against application failures, there are several options for future work.
In this paper, we assumed that the workload to be monitored runs entirely in one container inside \ac{K8s}.
However, the navigation stack for instance, consists of multiple individual workloads such as localization, path planning and trajectory execution that could run in separate containers following a more micro-service-oriented architecture.
At the same time, these workloads critically depend on each other in the sense that a failure in the localization module directly affects the functionality of the path planning module.
Hence, in a micro-service-oriented software architecture, the monitoring and failure mitigation system needs to take such dependencies into account and act accordingly when mitigating failures.
Although our proposed monitoring system is, in principle, able to encode such a hierarchical monitoring system in the substrate of Behaviour Trees, more research and evaluation is necessary to realize it.
Additionally, we aim to add monitoring vectors along the lines of our taxonomy as described in Sec.~\ref{ssubsec:taxonomy}, for instance, safety-focused monitoring.
Finally, we aim to incorporate our monitoring and failure mitigation system in a larger orchestration solution, which is not only able to mitigate failures but also reschedule containers and compute resources in case the monitoring solution detects sub-optimal behaviour of individual workloads above failure-level.

\addtolength{\textheight}{-2cm}   







\bibliographystyle{IEEEtran}
\bibliography{IEEEabrv, literature}

\end{document}